\documentclass[11pt]{article}

\usepackage{amsmath, amssymb, graphicx, cite}
\usepackage{hyperref}
\usepackage{caption}
\usepackage{subcaption}
\usepackage{geometry}
\usepackage{colortbl}
\usepackage{algorithm}
\usepackage{algpseudocode}
\usepackage{subcaption}
\usepackage{float}
\usepackage[table]{xcolor}
\usepackage{fancyhdr}

\definecolor{lightgray}{gray}{0.8}  
\definecolor{darkgray}{gray}{0.6} 

\title{SOLD: SELFIES-based Objective-driven Latent Diffusion}
\author{Elbert Ho \\ The Pingry School}
\date{}

\begin{document}
\linespread{1.5}
\maketitle
\thispagestyle{empty}
\newpage

\thispagestyle{empty}
\begin{abstract}
    Recently, machine learning has made a significant impact on 
    \textit{de novo} drug design. However, current approaches to creating 
    novel molecules conditioned on a target protein typically rely on generating 
    molecules directly in the 3D conformational space, which are often slow and overly complex.
    In this work, we propose SOLD (SELFIES-based Objective-driven Latent Diffusion), a novel 
    latent diffusion model that generates molecules in a latent space 
    derived from 1D SELFIES strings and conditioned on a target protein. In the 
    process, we also train an innovative SELFIES transformer and propose a new way to 
    balance losses when training multi-task machine learning models.
    Our model generates high-affinity molecules for the target protein 
    in a simple and efficient way, while also leaving room for 
    future improvements through the addition of more data. 
\end{abstract}
\newpage
\setcounter{page}{1}
\section{Introduction}

In the past few years, major advancements have been made in
using machine learning for
\textit{de novo} drug design. Essentially, the goal is to generate novel 
molecules that are able to bind to a target protein with high affinity in order 
to inhibit its function, useful for developing new drugs. 
In the past, molecules were 
typically converted to one-dimensional (1D) SMILES strings and then fed into 
different models including LSTMs \cite{ERTL17}, VAEs \cite{Bombarelli16}, 
GANs \cite{Guimaraes18}, and GPTs \cite{Li23}. 
SMILES strings were later replaced with SELFIES strings to solve
the issue of generating invalid SMILES strings when the string does not follow the syntax of 
SMILES and cannot be turned into a molecule \cite{Krenn19}.
However, these 1D models are limited 
in that they do not capture all of the information in a molecule. 

Recent work has shown the effectiveness of three-dimensional (3D) models in generating 
novel molecules. Initial models focused on autoregressive ``growing"
models that generated 3D structures bit by bit \cite{gebauer20}. 
More recently, diffusion models have demonstrated superior performance 
in the domain of molecule generation. \cite{hoogeboom22}\cite{vignac2023midimixedgraph3d}\cite{schneuing2023}.
However, these models run in the
extremely high dimensional atomic space which makes training and 
convergence difficult. To resolve this, Xu et al. \cite{xu2023geometriclatentdiffusionmodels} proposed a 
method to train a diffusion model in the latent space of a VAE.
Still, this method is limited in that it is not designed to generate
ligands with a specific target protein in mind, unlike other
diffusion models that are conditioned on a target protein or binding
pocket. Furthermore, though 3D models can incorporate more information, they
rely on the use of complex equivariant graph neural networks (EGNNs)
which increase model complexity and hurt convergence. In addition, 
this limitation means that they are unable to adapt many recent
advancements in image generation models for use in drug discovery.

In this work, we propose SELFIES-based Objective-driven Latent Diffusion, or SOLD,
a novel 1D SELFIES latent diffusion model
that is able to generate novel molecules conditioned on a target
protein. We demonstrate that our model is able to generate novel ligands 
that are able to bind to a target protein with high affinity at a 
similar rate to other state-of-the-art approaches. Thus, although we may 
miss out on some 3D information, we find that the diffusion model is powerful 
enough to overcome this limitation. In other words, the model's 1D nature 
does not significantly impact its performance. 
To the best of our knowledge, this is the first work to successfully generate novel molecules 
using diffusion in the 1D space, and the first to generate
molecules conditioned on a target protein using a latent diffusion 
model.

Our work is unique in that it is able to generate novel molecules 
simply and efficiently compared to other state-of-the-art (SOTA) models because of its 1D latent space. 
Additionally, our model is much more scalable to larger datasets as it does not require the 
3D conformational 
information of the protein to be known due to 
a ligand-based-drug-discovery approach rather than a structure-based 
one.

\section{Related Work}

\textbf{Conditional Diffusion Models}. In the domain of 
image generation, recent works have focused 
on conditioning the model on either a target label or target text 
to generate realistic images that follow a prompt. In particular, 
the DALL-E models \cite{ramesh2021zeroshottexttoimagegeneration} \cite{ramesh2022hierarchicaltextconditionalimagegeneration} by OpenAI have been both popular and effective.
However, our approach follows the GLIDE architecture proposed by 
Nichol and Dhariwal \cite{nichol2022glidephotorealisticimagegeneration}, as it is most easily adapted to the 
molecular domain.

\textbf{Stable/Latent Generative Models}. Due to the high complexity 
of the original space in many generative models, recent works have 
attempted to train models in a lower-dimensional latent space \cite{kalwar2022latentganautoencoderlearningdisentangled} \cite{rombach2022highresolutionimagesynthesislatent}. These 
models are trained on a variety of domains including 
image generation and text generation. Most 
notably, Rombach et al. \cite{rombach2022highresolutionimagesynthesislatent}. trained an image diffusion model in the 
latent space known as Stable Diffusion that has performed extremely 
well in image generation tasks. For molecule generation, Xu et al. \cite{xu2023geometriclatentdiffusionmodels} trained a 
diffusion model in the latent space of a VAE to generate novel molecules 
in the 3D space. Our study differs in that we generate molecules 
in the 1D space and condition on a target protein. We also do not use VAEs.
We choose to use a latent space 
because past works on 1D molecular generation have shown that using the SMILES space directly 
is difficult and not very effective. 

\section{Background}
\subsection{SELFIES}
SELFIES (Self-Referencing Embedded Strings) improve upon the SMILES (Simplified Molecular Input Line Entry System) 
representation of 
molecules \cite{Krenn19}. In the original SMILES method, molecules are first represented 
as a graph where atoms are nodes and bonds are edges. The graph is then turned into a 
spanning tree by removing cycles such as benzene rings. Finally, the tree is traversed in a 
DFS manner to generate the SMILES string. However, this method is not perfect as there can be 
invalid SMILES strings. SELFIES strings, on the other hand, use formal grammar to ensure that 
all generated strings are valid. In particular, the parser is able to keep track of the potential 
valid next symbols and simply skip over the invalid ones. This makes SELFIES strings 
robust and more useful for generating novel molecules.

\subsection{DDPMs}
Denoising diffusion probabilistic models (DDPMs) were first
introduced by Sohl-Dickstein et al. \cite{sohldickstein2015deepunsupervisedlearningusing}. Diffusion 
algorithms model a probabilistic Markov chain that follows the distribution below:
\begin{equation}
    q(z_t | z_{t - 1}) := \mathcal{N}(\sqrt{\alpha_t}z_{t - 1}, 1-\alpha_t \mathcal{I}) 
\end{equation}
A U-Net is used to model the reverse diffusion process 
represented by 
\begin{equation}
    p_{\theta}(z_{t-1} | z_t, y) := \mathcal{N}(\mu_{\theta}(z, t, y), \Sigma_{\theta}(z, t, y))
\end{equation}
where $y$ is the label or condition. The trained backward 
model is then used to generate samples by starting from 
$z_t \sim \mathcal{N}(0, \mathcal{I})$ and using 
$p_{\theta}(z_{t-1} | z_t, y)$ to go to $z_{t-1}$ and then $p_{\theta}(z_{t-2} | z_{t-1}, y)$ to go to $z_{t-2}$ and so on
until we reach $z_0$ which is our final sample.

In practice, the model is optimized according to the improved loss function
proposed by Nichol and Dhariwal \cite{nichol2021improveddenoisingdiffusionprobabilistic} below:
\begin{gather}
    L_{vlb} := L_0 + L_1 + \dots + L_T \\
    L_0 := -\log p_{\theta}(z_0 | y) \\
    L_{t-1} := D_{KL}(q(z_{t-1} | z_0) || p_{\theta}(z_{t-1} | z_t, y)) \\
    L_{T} := D_{KL}(q(z_{T} | z_0) || p(z_{T})) \\
    L_{simple} := \mathbb{E}_{t, z_0, \epsilon}[||\epsilon - \epsilon_{\theta}(z_t, t, y)||^2] \\
    L_{final} := L_{simple} + \lambda L_{vlb}
\end{gather}

Essentially, the simple loss is the mean squared error between the predicted 
noise and the actual noises. The VLB loss, on the other hand, is the sum of the 
KL divergences between the predicted distribution and the true distribution at each time 
step. $L_0$ is defined as the negative log likelihood of the distribution. It 
is used to help the model learn the standard deviation of the distribution rather than just the mean.
$\lambda$ is a hyperparameter that controls the weight of the VLB loss, and it is 
set to a low value (.0001) to prevent it from dominating the loss.
When sampling, $\mu_{\theta}$ is derived from $\epsilon_{\theta}$ \cite{ho2020denoisingdiffusionprobabilisticmodels}.

\subsection{Classifier-Free Guidance}
Classifier-free guidance is a simple method proposed by
Ho \& Salimans \cite{ho2022classifierfreediffusionguidance} to improve the performance
of conditional diffusion models. The idea is to train the model to 
recognize the general distribution of all possible samples by
occasionally replacing $y$ with $\emptyset$ when training the model.
When sampling, we use 
\begin{equation}
    \tilde{\epsilon}_{\theta}(z, t, y) = (1 + w)\epsilon_{\theta}(z, t, y)
    - w\epsilon_{\theta}(z, t, \emptyset)
\end{equation}
so that the final sample is closer to the conditioned distribution
than to the unconditional distribution. Increasing $w$ increases the 
guidance. 

\begin{figure*}[ht]
    \centering
    \includegraphics[width=\textwidth]{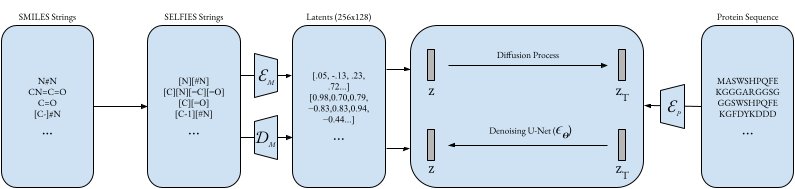}
    \caption{Illustration of SOLD. 
    Molecules are first 
    encoded as SELFIES strings and then transformed into the 
    latent space through a transformer. A diffusion model 
    is then trained in the latent space to generate novel
    molecules. Proteins are encoded as ESM-2 embeddings on the 
    right.}
\end{figure*}

\section{Methods}
Inspired by stable diffusion models, we train and use a latent
diffusion model to generate novel molecules. Figure 1 \footnote{All figures and images in 
this paper were created by student researcher} is a graphical 
pipeline of the model, showing how molecules are encoded, decoded, and then 
generated. We first represent
molecules as SELFIES strings and use a transformer to learn a 
latent representation of the molecule in Section 4.1. We then
detail the specifics of the diffusion model in Section 4.2. 
\subsection{Molecular Encoder}
\textbf{Model Architecture}.
Since transformers \cite{vaswani2023attentionneed} have been shown to perform extremely 
well on NLP tasks \cite{brown2020languagemodelsfewshotlearners} and have also been shown to be effective 
when applied to SMILES strings \cite{honda2019smilestransformerpretrainedmolecular} \cite{fabian2020molecularrepresentationlearninglanguage}, we use a transformer to learn 
representations of SELFIES strings. We have attempted to use a VAE 
instead to exploit the inherently probabilistic nature of its latent space but found that the
VAE was unable to learn a sufficiently accurate enough latent 
representation (our experiments showed that 0\% of sequences were correctly reconstructed). In particular, 
while VAEs may be able to learn an approximation of a latent space, they are 
fundamentally flawed for our purposes as they are not able to reconstruct inputs very well due to their 
stochastic nature. Thus, our VAE begins to act as a generator rather than just an encoder and decoder, which could 
interfere with our diffusion model. 

In comparison, transformers can learn an effective reconstruction of the latent space 
because their attention mechanism allows them to learn the relationships between different parts 
of the SELFIES string and thus generate a more accurate latent representation. Our model has 2 transformer 
layers each with embedding dimension 256 and 4-heads of attention.

\textbf{Training}.
We sourced 10,000 molecules from the CHEMBL dataset \cite{zdrazil23}.
First, molecules were converted to SMILES and then randomized 
to avoid biasing the model towards the canonical SMILES sourced from 
CHEMBL \cite{Arus-Pous2019}. Then, molecules were converted to SELFIES strings 
because SELFIES strings are guaranteed to be valid, something 
that is particularly useful when generating new molecules \cite{Krenn19}. 
SELFIES strings were first tokenized using the standard SELFIES tokenizer. 
We then trained and used a byte pair encoder (BPE)
\cite{sennrich2016neuralmachinetranslationrare} as we needed to 
trade-off latent space size for vocabulary size. This is because we will be 
training the diffusion model on the latent space and we would rather use
a smaller latent space later on. In particular, we tried to train without
byte pair encoding but this forced us to either truncate the SELFIES strings 
(which would be problematic as the generative model would have to understand incomplete molecules)
or use a much larger latent space to fit all of the SELFIES strings. Our BPE model had a vocabulary size of 
256 as we found this to reduce most of the SELFIES strings to a 
length less than 128. All SELFIES strings were then tokenized using the 
SELFIES tokenizer and BPE tokenizer before being padded to length 128 or simply 
removed from the dataset if they had a length longer than 128. 

As suggested by Honda et al. \cite{honda2019smilestransformerpretrainedmolecular}, we train the model in a 
multi-task setting where the final layer of the transformer 
is used to predict the SPS (Spacial Score), MinEStateIndex 
(electrotopological state index), ExactMolWt (exact molecular
weight), BalabanJ (Balaban J index), and VSA\_EState6 (EState
fragment contribution to the sixth bin of the VSA\_EState). 
These tasks were chosen using PCA analysis on the available
RDKit \cite{rdkit} molecular descriptors. 

In essence, given an encoder $\mathcal{E}$ that maps SELFIES tokens
to a latent representation and a decoder $\mathcal{D}$ that maps
latent representations to SELFIES tokens, we train the model
to minimize the standard cross-entropy reconstruction loss 
$\ell(\mathcal{D}(\mathcal{E}(x)), x)$ as well as task-specific 
losses $\ell(\mathcal{D}_t(\mathcal{E}(x)), y_t)$ where $y_t$ is the
target value for task $t$ (all mean squared error). 

To optimize, we 
modify the multi-task loss training approach proposed by Lin et al.
\cite{lin2023dualbalancingmultitasklearning},
which uses a momentum-based approach combined with gradient normalization,
by integrating the second moment found in the Adam optimizer \cite{kingma2017adammethodstochasticoptimization}. We found that adding the second moment significantly improved convergence. 

We outline our proposed novel training function below in Algorithm 1, where 
the second moment is represented by $\nu$ in lines 8 and 16. 

\begin{algorithm}[H]
    \caption{Dual-Balancing Multi-Task Learning with Adam Modifications}
    \label{alg:dual_balancing}
    \begin{algorithmic}[1]
    \Require numbers of iterations $T$, learning rate $\eta$, tasks $\{ \mathcal{D}_k \}_{k=1}^K$, $\beta$, $\beta_2$, $\epsilon = 10^{-8}$
    \State randomly initialize $\theta_0$, $\{\psi_{k,0}\}_{k=1}^K$
    \State initialize $\hat{g}_{t,-1} = 0$, for all $t$
    \For{$t = 0, \ldots, T-1$}
        \For{$k = 1, \ldots, K$}
            \State sample a mini-batch dataset $\mathcal{B}_{k,t}$ from $\mathcal{D}_k$
            \State $g_{k, t} = \nabla_{\theta_t} \log(\ell_k(\mathcal{B}_{k, t}; \theta_t, \psi_{k, t}) + \epsilon)$
            \State compute $\mu_{k, t} = \frac{\beta \mu_{k, t-1} + (1 - \beta) g_{k, t}}{1 - \beta^t}$
            \State compute $\nu_{k, t} = \frac{\beta_2 \nu_{k, t - 1} + (1 - \beta_2) g_{k, t}^2}{1 - \beta_{2}^{t}}$
            \State $\hat{g}_{k, t} = \frac{\mu_{k, t}}{\sqrt{\nu_{k, t}} + \epsilon}$
        \EndFor
        \State compute $\tilde{g}_t = \alpha_t \sum_{k=1}^K \frac{\hat{g}_{k, t}}{\|\hat{g}_{k, t}\|_2 + \epsilon}$, where $\alpha_t = \max_{1 \leq k \leq K} \|\hat{g}_{k, t}\|_2$
        \State update task-sharing parameter by $\theta_{k+1} = \theta_k - \eta \tilde{g}_k$
        \For{$k = 1, \ldots, K$}
            \State $g'_{k, t} = \nabla_{\theta_t} \log(\ell_k(\mathcal{B}_{k, t}; \theta_t, \psi_{k, t}) + \epsilon)$
            \State compute $\mu'_{k, t} = \frac{\beta \mu'_{k, t-1} + (1 - \beta) g'_{k, t}}{1 - \beta^t}$
            \State compute $\nu'_{k, t} = \frac{\beta_2 \nu'_{k, t - 1} + (1 - \beta_2) g_{k, t}^{'2}}{1 - \beta_{2}^{t}}$
            \State $\psi_{k,t+1} = \psi_{k, t} - \eta \frac{\mu'_{k, t}}{\sqrt{\nu'_{k, t}} + \epsilon}$
        \EndFor
    \EndFor
    \Return $\theta_T$, $\{\psi_{k,T}\}_{k=1}^K$
    \end{algorithmic}
\end{algorithm}

The model was trained for a total of 200 epochs with 20 epochs 
used for pretraining without the task losses. This was done to 
help the model first learn a robust latent representation before 
simultaneously learning the tasks. Cosine annealing was used for the learning rate 
which was set to start at .0001.

\subsection{Latent Diffusion Model}
\textbf{Model Architecture}.
Once the SELFIES strings are encoded into the latent space, we 
train a diffusion model to generate novel molecules. We take the 
last layer of the transformer as our encoder $\mathcal{E}_M$ which
makes the latents essentially length 128 1D vectors that have 
256 channels. Before inputting into the model, we normalize the 
latents to have a mean of 0 and a standard deviation of 1 (using a global mean and standard deviation) 
before converting 
to 0 to 1 using the standard normal cumulative distribution function. In other words,
we calculate $\Phi(z) = \frac{1}{\sqrt{2 \pi}}$ $\int_{-\infty}^{z} e^{-\frac{t^2}{2}} dt$
which outputs in the range 0 to 1.
Finally, the input is linearly scaled to the range of -1 to 1. 
When going back to our original space, we reverse the process by applying the inverse normal cumulative distribution function and
truncate the final values to be between -6.5 and 6.5 because of precision.
We imitate the architecture proposed by Nichol and 
Dhariwal \cite{nichol2022glidephotorealisticimagegeneration} with the exception that we use a 1D U-Net instead 
of a 2D one. We used 1000 timesteps for the diffusion process.
In addition, instead of using text embeddings from a 
transformer as the conditioning for the model, we use ESM-2 \cite{Lin2022.07.20.500902} with 
embedding dimension of 1280 to embed proteins. 
ESM-2 is a transformer-based model 
that can be used to turn the sequence of amino acids into an embedding.
The ESM-2 model's weights are trained during the diffusion process.

\textbf{Training}.
Our model has 360M parameters and was trained on a dataset of 
15,000 protein-ligand pairs sourced from the PDBbind dataset \cite{Wang2005Jun}.
The model was trained for a total of 1000 epochs with batch size 
of 32 with 16-bit precision on an NVIDIA RTX 4090. Additionally, we 
trained with classifier-free guidance so 20\% of inputs were conditioned on 
$\emptyset$.

\textbf{Sampling}.
To sample from the model, we start with a random latent vector 
with dimensions of 256 x 128. We then use the reverse diffusion model
combined with classifier-free guidance to generate a sample. 
We found experimentally that a guidance weight of 5 worked 
the best. Latent samples were converted back to SELFIES strings 
through the decoder $\mathcal{D}_M$ and then converted to SMILES
strings which could then be converted to either 2D images or 
3D structures via RDKit.

\textbf{Optimizing Qualities}. 
To optimize the quality of generated molecules, we attempted to adopt the
simple evolutionary algorithm proposed by Schneuing et al. 
\cite{schneuing2023}. In particular, given a molecule with 
a desired binding affinity, we noise the latent representation 
by 75 steps before denoising it back to a molecule. In this 
way, we obtain similar molecules with slightly different properties.
By selecting the best molecules with desired properties (such as QED 
and SAS), we can once again noise these molecules and then denoise them
to obtain even better molecules. This process is repeated until 
the desired properties are met. We chose to noise by 75 steps because 
we found that this was the threshold where the molecules would be 
similar to the original but also have some diversity (Figure 2). Note that in Figure 2b, 
we use raw SA scores so lower is better and the range is 1 to 9.
Here, rather than using $w = 5$ for the guidance weight 
as before, we lower it to $w = 0$, as increasing $w$ was found to lower diversity 
which is not desirable in this case. Though, as discussed later, we found that 
the evolutionary algorithm did not significantly improve the quality of the molecules 
and actually hurt the Vina score considerably.

\begin{figure}[H]
    \centering
    \resizebox{1\textwidth}{!}{ 
        \begin{subfigure}[b]{0.45\textwidth}
            \centering
            \includegraphics[width=\textwidth]{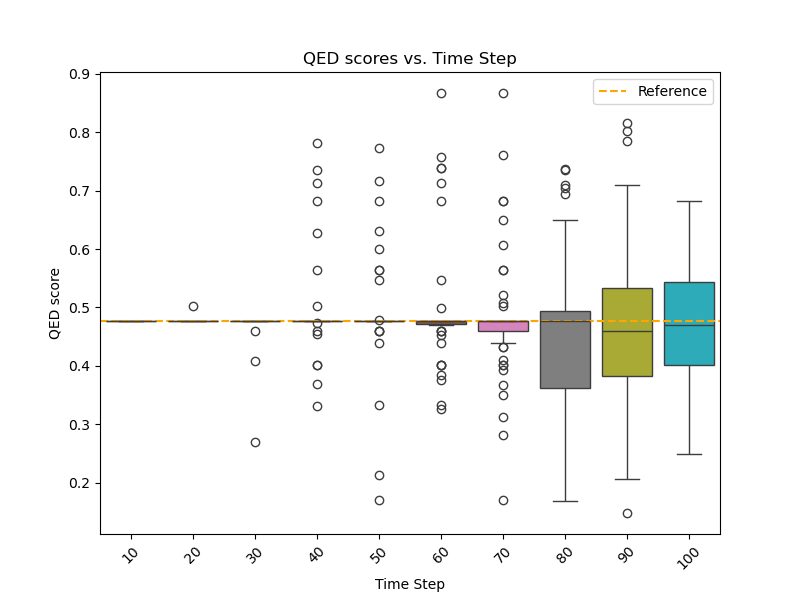}
            \caption{QED Scores when noising and denoising from 0 to 100 steps}
            \label{fig:sub1a}
        \end{subfigure}
        \hfill
        \begin{subfigure}[b]{0.45\textwidth}
            \centering
            \includegraphics[width=\textwidth]{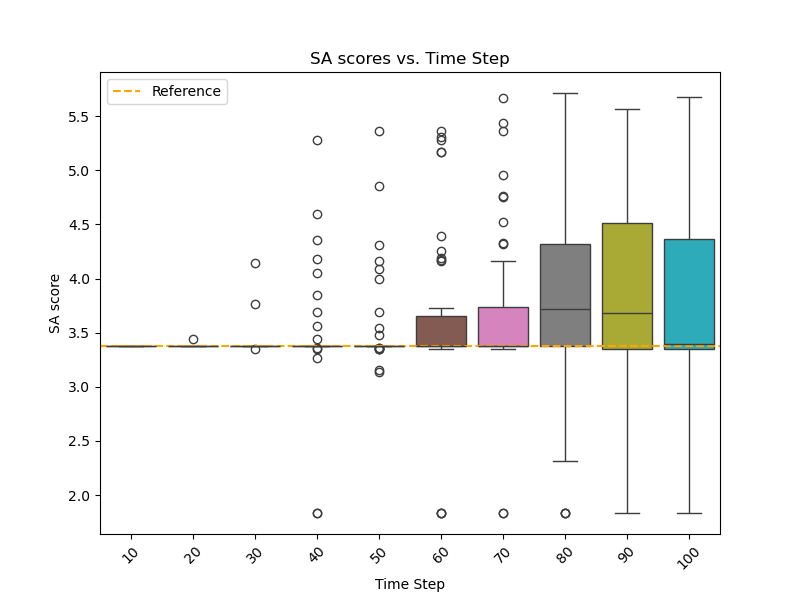}
            \caption{SAS Scores when noising and denoising from 0 to 100 steps}
            \label{fig:sub2}
        \end{subfigure}
        \hfill
        \begin{subfigure}[b]{0.45\textwidth}
            \centering
            \includegraphics[width=\textwidth]{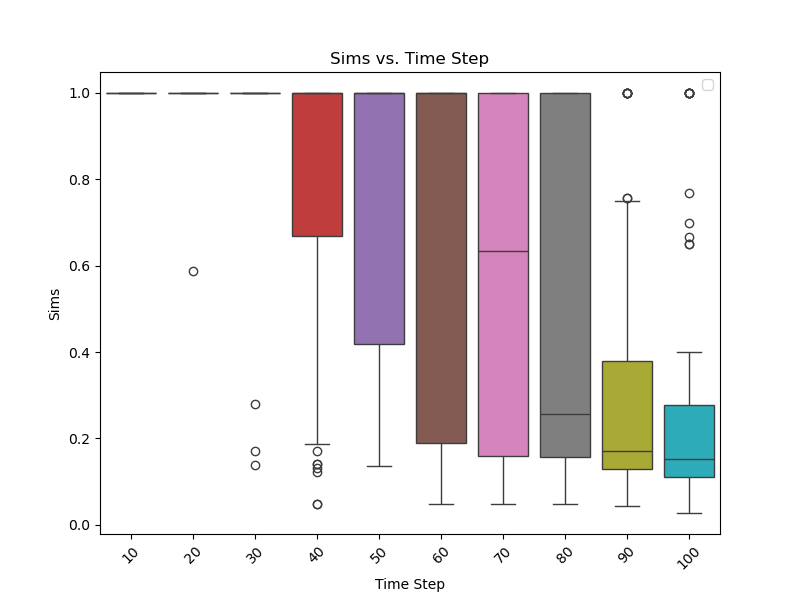}
            \caption{Similarity to original molecule when noising and denoising from 0 to 100 steps}
            \label{fig:sub3}
        \end{subfigure}
    }
    \caption{Some results of the evolutionary algorithm.}
    \label{fig:main}
\end{figure}

\section{Experiments}

\textbf{Baselines}. We compare our model to the following SOTA 
baselines. \textbf{3D-SBDD} \cite{luo20223dgenerativemodelstructurebased}
was one of the first 3D conditional molecule generative models, 
and it uses an autoregressive approach to sample the 3D coordinates of 
each atom in a molecule
from a predicted probability distribution. \textbf{Pocket2Mol} \cite{peng2022pocket2molefficientmolecularsampling}
is a similar approach except it takes into consideration bond types 
and functional groups. \textbf{TargetDiff} \cite{guan2023d} and 
\textbf{DiffSBDD-cond} \cite{schneuing2023} are similar 
3D diffusion models that are conditioned on a target protein pocket. There were 
also a few models that we looked at such as DrugGPT which is a relatively 
recent 1D model that uses a GPT-like architecture rather than diffusion
to generate molecules.
However, since none of their papers reported Vina score as a metric and we 
were unable to run their model with our limited resources, we do not include them 
in the comparisons below.

\textbf{Metrics}. We evaluate our model on the following metrics.
\begin{itemize}
\item \textbf{Vina} is the binding energy of the generated 
ligand with the target protein or protein-pocket as predicted by 
AutoDock Vina. We used UniDock \cite{Yu23} (a GPU accelerated version of AutoDock)
for our experiments, but the results are comparable. Additionally, we used the 
standard parameters (exhaustiveness: 128, max step: 20, num modes: 3, refine step: 3, top n: 100).
Due to 
compute constraints, we only evaluate our model on a few select 
proteins not in the training set and take the average. In particular,
we show results for our tests with the COVID-19 3C-like main protease ($3CL\textsuperscript{pro}$), which 
is a known target for COVID-19 drugs \cite{Zhu2020-be}.
\item \textbf{QED} \cite{Bickerton2012-vd}
(quantitative estimate of drug-likeness) is a metric that measures 
how ``drug-like" a molecule is based on a weighted sum of different 
molecular properties such as logP and molecular weight. 
\item \textbf{SA}
\cite{Ertl2009}
(synthetic accessibility score) calculates how "easy" it would be 
to synthesize the molecule based on composite fragments. Here, scores 
are normalized between 0 and 1 and reversed so that higher scores 
are better.

\item \textbf{Diversity} calculate the difference between generated molecules 
using the Tanimoto similarity between Morgan fingerprints \cite{Bajusz2015Dec}.

\item \textbf{Time} is the time in seconds taken to generate 100 molecules for a given 
protein or protein-pocket. Since this time is dependent on the GPU, we scale 
the time based on the GPU used. In particular, the V100 is close in speed to the 
RTX 3060 that we used for inference, but the A100 used for DiffSBDD is about 8 times 
faster. Thus, we scale their time up by 8.
\end{itemize}

\begin{table}[H]
    \centering
    \begin{tabular}{c|c|c|c|c|c}
    \hline
    \textbf{w} & \textbf{\# Filter ($\uparrow$)} & \textbf{Vina ($\downarrow$)} & \textbf{Vina (Top-10\%) ($\downarrow$)} & \textbf{Filtered ($\downarrow$)} & \textbf{Filtered (Top-10\%) ($\downarrow$)} \\
    \hline
    0 & \cellcolor{darkgray} 16 & -4.103 & -6.771 & -4.282 & -5.957 \\
    1 & 9  & -4.416 & -7.709 & \cellcolor{darkgray} -5.281 & \cellcolor{darkgray} -8.197 \\
    3 & 7  & \cellcolor{lightgray} -4.486 & -7.650 & \cellcolor{lightgray} -5.193 & -6.731 \\
    5 & \cellcolor{lightgray} 13 & \cellcolor{darkgray} -4.498 & \cellcolor{darkgray} -9.830 & -4.900 & -6.613 \\
    7 & 12 & -4.558 & \cellcolor{lightgray} -8.670 & -4.257 & -5.982 \\
    9 & 11 & -3.415 & -6.600 & -5.133 & -6.875 \\
    11 & 12 & -2.935 & -6.429 & -4.609 & \cellcolor{lightgray}-7.131 \\
    \hline
    \end{tabular}
    \caption{Summary of Results for Different Values of $w$.}
\end{table}

First, we evaluate the effect of the guidance weight $w$ on the 
performance of our model (Table 1). We do this by testing $w$ from 0 to 11 
and evaluating the Vina score. We report two different metrics: the first is the entire set of generated molecules, and
the second is 
a "filtered" list where we only take molecules of QED at least 0.40 
and SAS at least 0.33. We find that $w = 5$ performs the best in terms of 
Vina score and top 10\% for the overall set of molecules. Note that the 
filtered metrics have large uncertainty due to small sample size, so we choose 
to rely more on the overall Vina metrics. Therefore, we use $w = 5$ for all remaining 
experiments.

\begin{table}[H]
    \centering
    \begin{tabular}{l|c|c|c|c|c|c}
    \hline
    \textbf{Model} & \textbf{Vina ($\downarrow$)} & \textbf{Vina (Top-10\%) ($\downarrow$)} & \textbf{QED ($\uparrow$)} & \textbf{SA ($\uparrow$)} & \textbf{Div. ($\uparrow$)} & \textbf{Time ($\downarrow$)} \\ \hline
    3D-SBDD \cite{luo20223dgenerativemodelstructurebased} & -5.888 & -7.289 & \cellcolor{lightgray}0.502 & \cellcolor{lightgray}0.675 & \cellcolor{lightgray}0.742 & 19659 \\ 
    Pocket2Mol \cite{peng2022pocket2molefficientmolecularsampling} & -7.058 & -8.712 & \cellcolor{darkgray}\textbf{0.572} & \cellcolor{darkgray}\textbf{0.752} & 0.735 & 2504 \\ 
    TargetDiff \cite{guan2023d} & \cellcolor{lightgray}-7.318 & -9.669 & 0.483 & 0.584 & 0.718 & 3428 \\ 
    DiffSBBDD \cite{schneuing2023} & \cellcolor{darkgray}\textbf{-7.333} & \cellcolor{darkgray}\textbf{-9.927} & 0.475 & 0.612 & 0.725 & \cellcolor{lightgray}1088 \\ 
    \hline 
    SOLD (Ours) & -4.498 & \cellcolor{lightgray}-9.830 & 0.3727 & 0.451 & \cellcolor{darkgray}\textbf{0.946} & \cellcolor{darkgray}\textbf{960} \\ \hline
    \end{tabular}
    \caption{Comparison of SOTA models for molecular generation tasks.}
\end{table}

We find that our model performs comparably to other existing models (Table 2). 
In particular, though our average Vina score is slightly lower than other 
models, our top 10\% Vina score is comparable. Since drug discovery 
primarily depends on the best hits, we believe that the weakness in 
the average Vina score is not a major issue. Additionally, our model 
is able to generate molecules faster than other SOTA models. Combined with 
our higher diversity, we are able to therefore quickly explore a much larger 
chemical space. As such, we believe that using SMILES strings 
as an input can be advantageous because it is much easier to explore 
a larger chemical space with 1D models than 3D models. 

Below, we show two generated molecules for the COVID-19 $3CL\textsuperscript{pro}$. In Figures 3 and 4, we show 
the 2D rendering of the generated molecule as well as its location in the 
larger protein and in the binding pocket. We also show the hydrophobic 
surface of the binding pocket with the bound ligand. These figures were generated by 
using the Chimera software. For comparison, Figure 5 
shows the known drug Nirmatrelvir, which is a part of the Paxlovid drug combination 
\cite{Reina2022-xy}. As can be seen, our model is able to generate molecules that 
have similar QED and SAS scores to known drugs but much better docking scores.

In particular, our docking score is intended to be a relative estimate of $\Delta G$. 
However, since Autodock Vina does not include the significant entropy increase from 
the ejection of water molecules from the binding site, it will be off by a constant factor. 
Here, the experimental $\Delta G$ of Nirmatrelvir is about -11.5 \cite{biom13091339} 
so we also subtract 5.60 from the 
$\Delta G$ of our compound (since our Vina output for Nirmatrelvir was -5.90).
We have that IC50 (half-maximal inhibitory concentration) is approximately equivalent to 
$K_i$ and we can relate $K_i$ to the 
true $\Delta G$ of the binding. We have that $IC_{50} \approx K_i = e^{\frac{\Delta G}{RT}}$ 
since $K_i = \frac{1}{K_d}$ and $K_d = e^{-\frac{\Delta G}{RT}}$
where $R$ is the gas constant and $T$ is the temperature. Calculating this value out 
for the known drug Paxlovid ($\Delta G \approx -11.5$) and our first molecule ($\Delta G \approx -13.4$), we get that the 
$IC_{50}$ of Paxlovid is approximately 7.8 nM and our first molecule 
is approximately 0.37nM. Thus, our model is able to generate molecules that are 
more than 20 times more potent than already effective drugs. 

\begin{figure}[H]
    \centering
    \resizebox{0.8\textwidth}{!}{ 
        \begin{tabular}{cc}
            \begin{subfigure}[b]{0.45\textwidth}
                \centering
                \includegraphics[width=\textwidth]{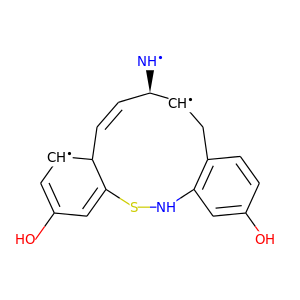}
                \caption{Illustration of one generated molecule. This 
                molecule had a QED of 0.505 and an SAS of 0.374 and an 
                affinity score of -7.70.}
                \label{fig:sub1b}
            \end{subfigure}
            &
            \begin{subfigure}[b]{0.45\textwidth}
                \centering
                \includegraphics[width=\textwidth]{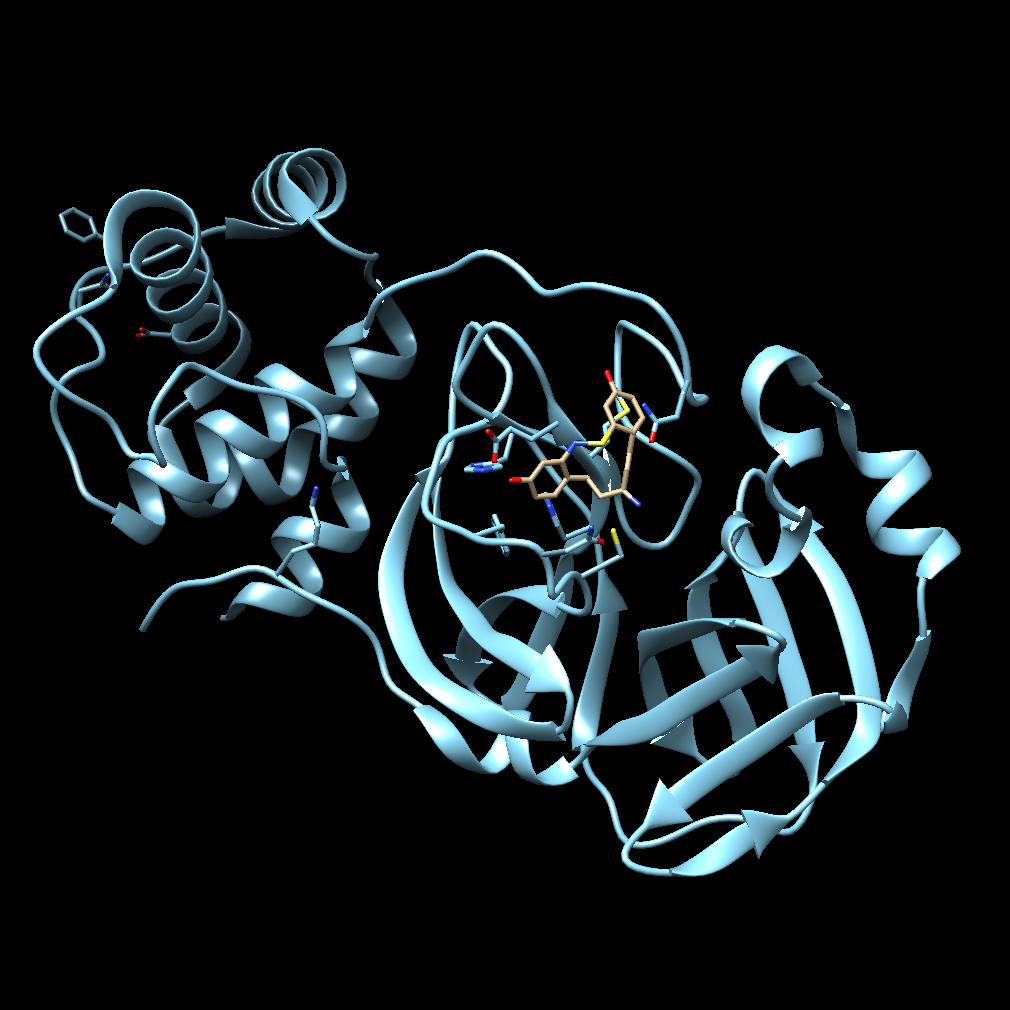}
                \caption{Rendering of molecule bound to $3CL\textsuperscript{pro}$ protease}
                \label{fig:sub2b}
            \end{subfigure}
            \\
            \begin{subfigure}[b]{0.45\textwidth}
                \centering
                \includegraphics[width=\textwidth]{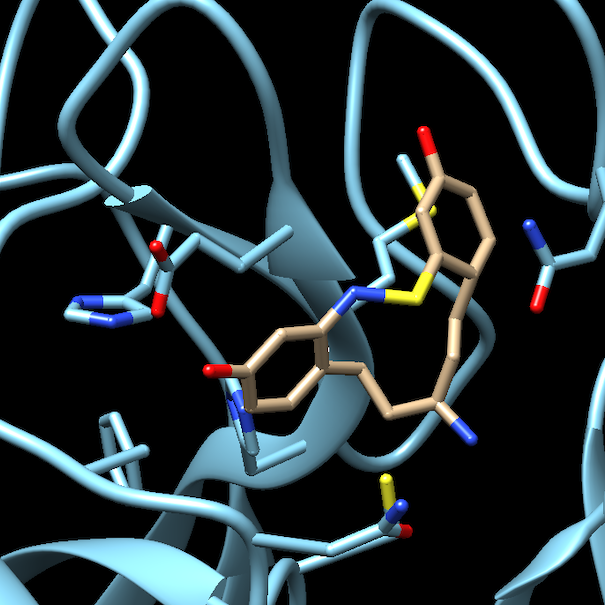}
                \caption{Rendering of molecule with $3CL\textsuperscript{pro}$ protease binding site}
                \label{fig:sub3b}
            \end{subfigure}
            &
            \begin{subfigure}[b]{0.45\textwidth}
                \centering
                \includegraphics[width=\textwidth]{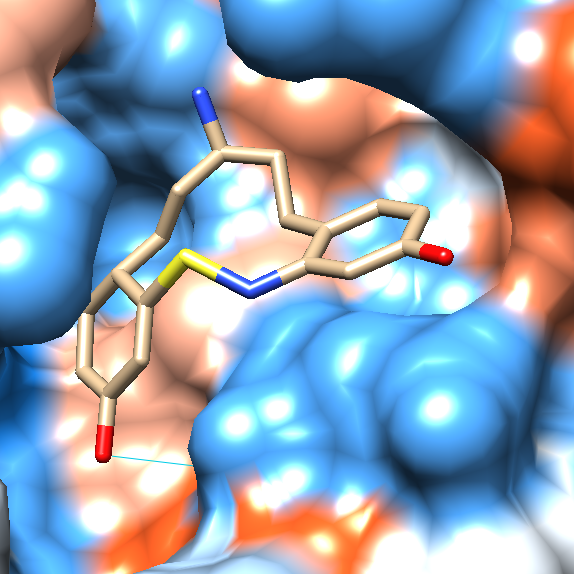}
                \caption{Hydrophobic surface view of binding site with bound ligand}
                \label{fig:sub4b}
            \end{subfigure}
        \end{tabular}
    }
    \caption{Renderings of molecule generated by SOLD}
    \label{fig:mainb}
\end{figure}

\begin{figure}[H]
    \centering
    \resizebox{0.8\textwidth}{!}{ 
        \begin{tabular}{cc}
            \begin{subfigure}[b]{0.45\textwidth}
                \centering
                \includegraphics[width=\textwidth]{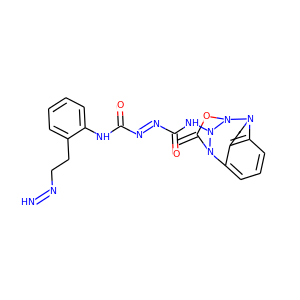}
                \caption{Illustration of one generated molecule. This 
                molecule had a QED of 0.594 and an SAS of 0.363 and an 
                affinity score of -7.15.}
                \label{fig:sub1d}
            \end{subfigure}
            &
            \begin{subfigure}[b]{0.45\textwidth}
                \centering
                \includegraphics[width=\textwidth]{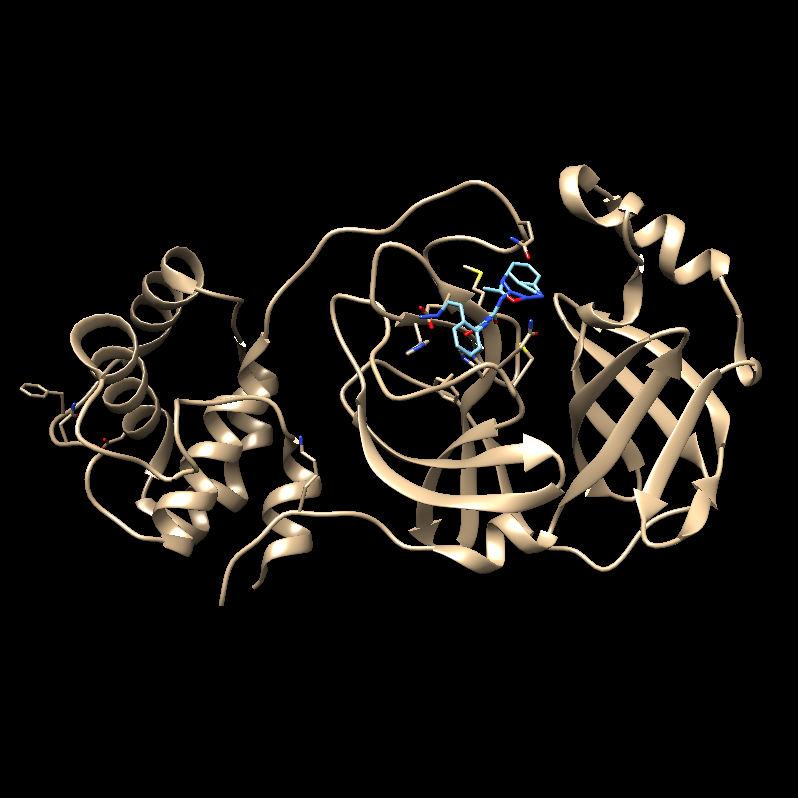}
                \caption{Rendering of molecule bound to $3CL\textsuperscript{pro}$ protease}
                \label{fig:sub2d}
            \end{subfigure}
            \\
            \begin{subfigure}[b]{0.45\textwidth}
                \centering
                \includegraphics[width=\textwidth]{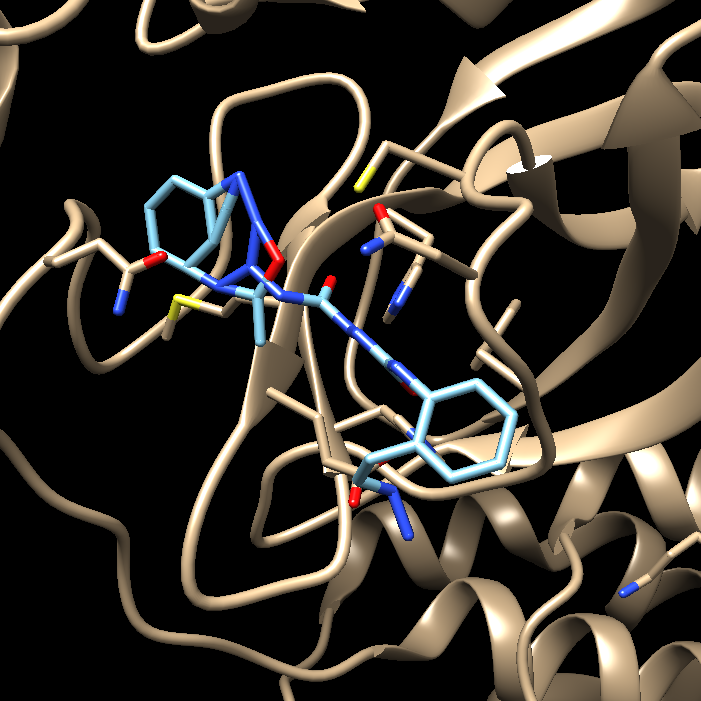}
                \caption{Rendering of molecule with $3CL\textsuperscript{pro}$ protease binding site}
                \label{fig:sub3d}
            \end{subfigure}
            &
            \begin{subfigure}[b]{0.45\textwidth}
                \centering
                \includegraphics[width=\textwidth]{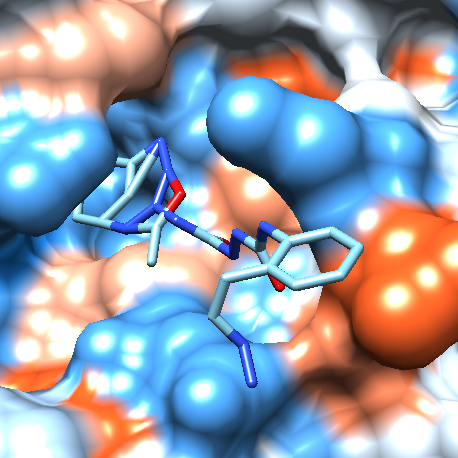}
                \caption{Hydrophobic surface view of binding site with bound ligand}
                \label{fig:sub4d}
            \end{subfigure}
        \end{tabular}
    }
    \caption{Render of another molecule generated by SOLD}
    \label{fig:maind}
\end{figure}

\begin{figure}[H]
    \centering
    \resizebox{0.8\textwidth}{!}{ 
        \begin{tabular}{cc}
            \begin{subfigure}[b]{0.45\textwidth}
                \centering
                \includegraphics[width=\textwidth]{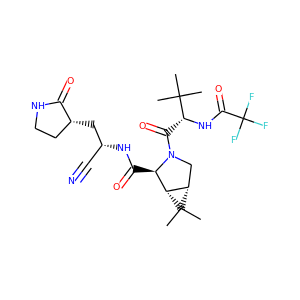}
                \caption{Illustration of Paxlovid (Nirmatrelvir). This 
                molecule had a QED of 0.504 and an SAS of 0.491 and an 
                affinity score of -5.90.}
                \label{fig:sub1c}
            \end{subfigure}
            &
            \begin{subfigure}[b]{0.45\textwidth}
                \centering
                \includegraphics[width=\textwidth]{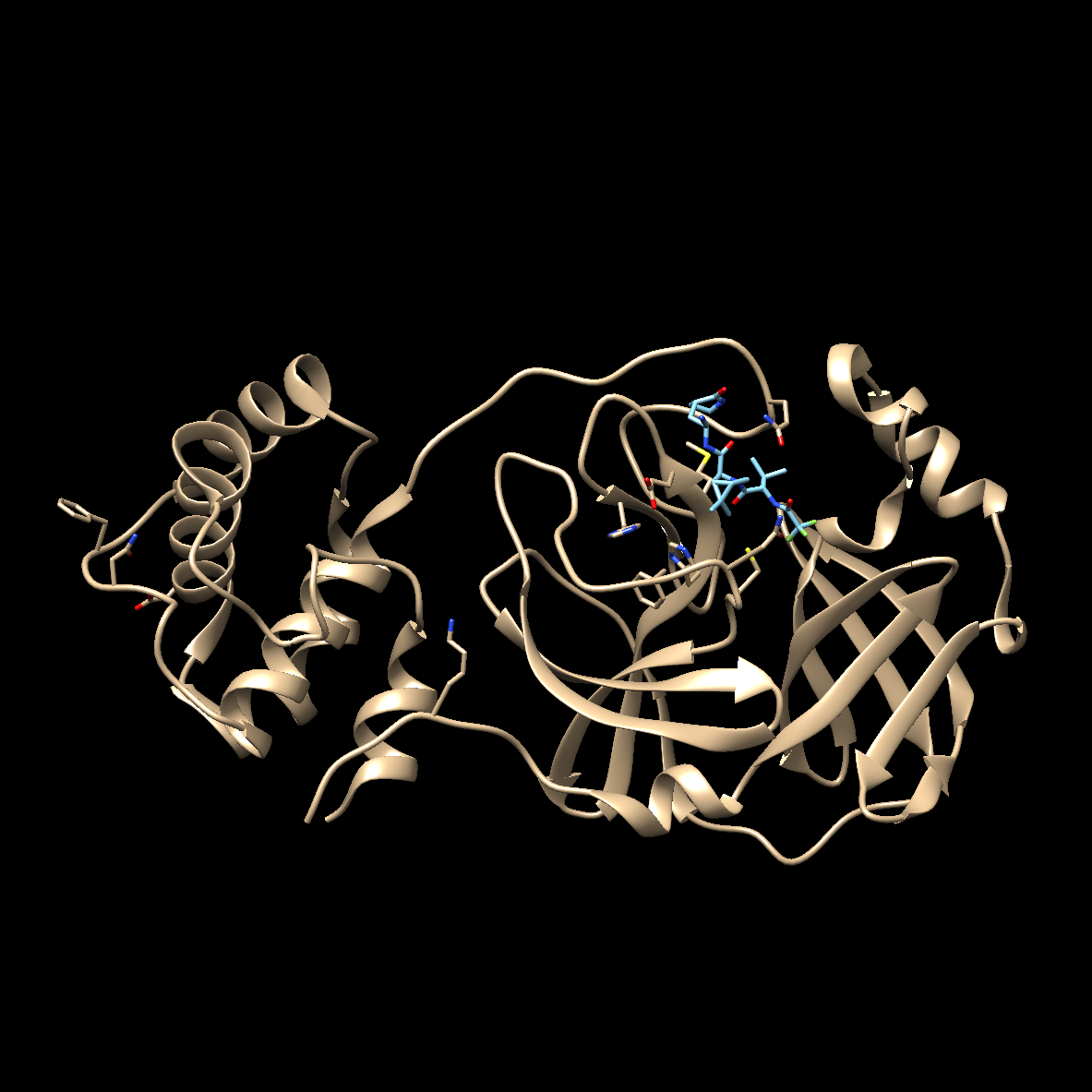}
                \caption{Rendering of molecule bound to $3CL\textsuperscript{pro}$ protease}
                \label{fig:sub2c}
            \end{subfigure}
            \\
            \begin{subfigure}[b]{0.45\textwidth}
                \centering
                \includegraphics[width=\textwidth]{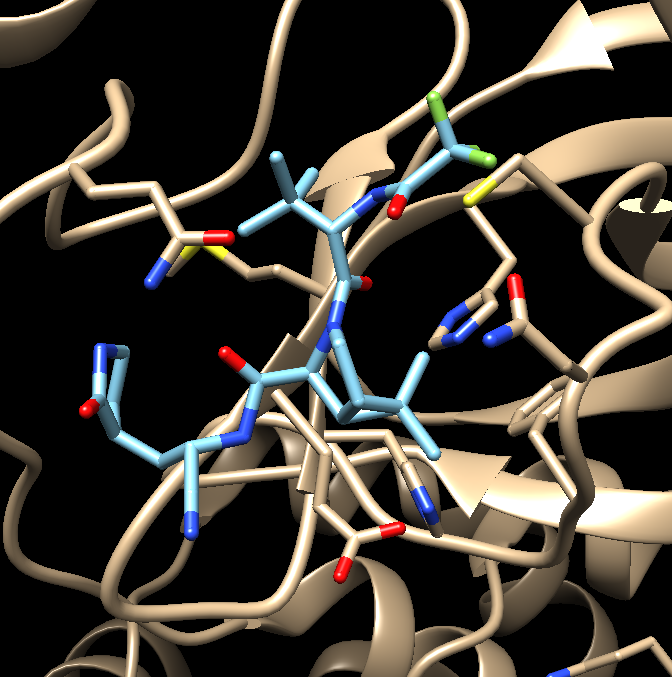}
                \caption{Rendering of molecule with $3CL\textsuperscript{pro}$ protease binding site}
                \label{fig:sub3c}
            \end{subfigure}
            &
            \begin{subfigure}[b]{0.45\textwidth}
                \centering
                \includegraphics[width=\textwidth]{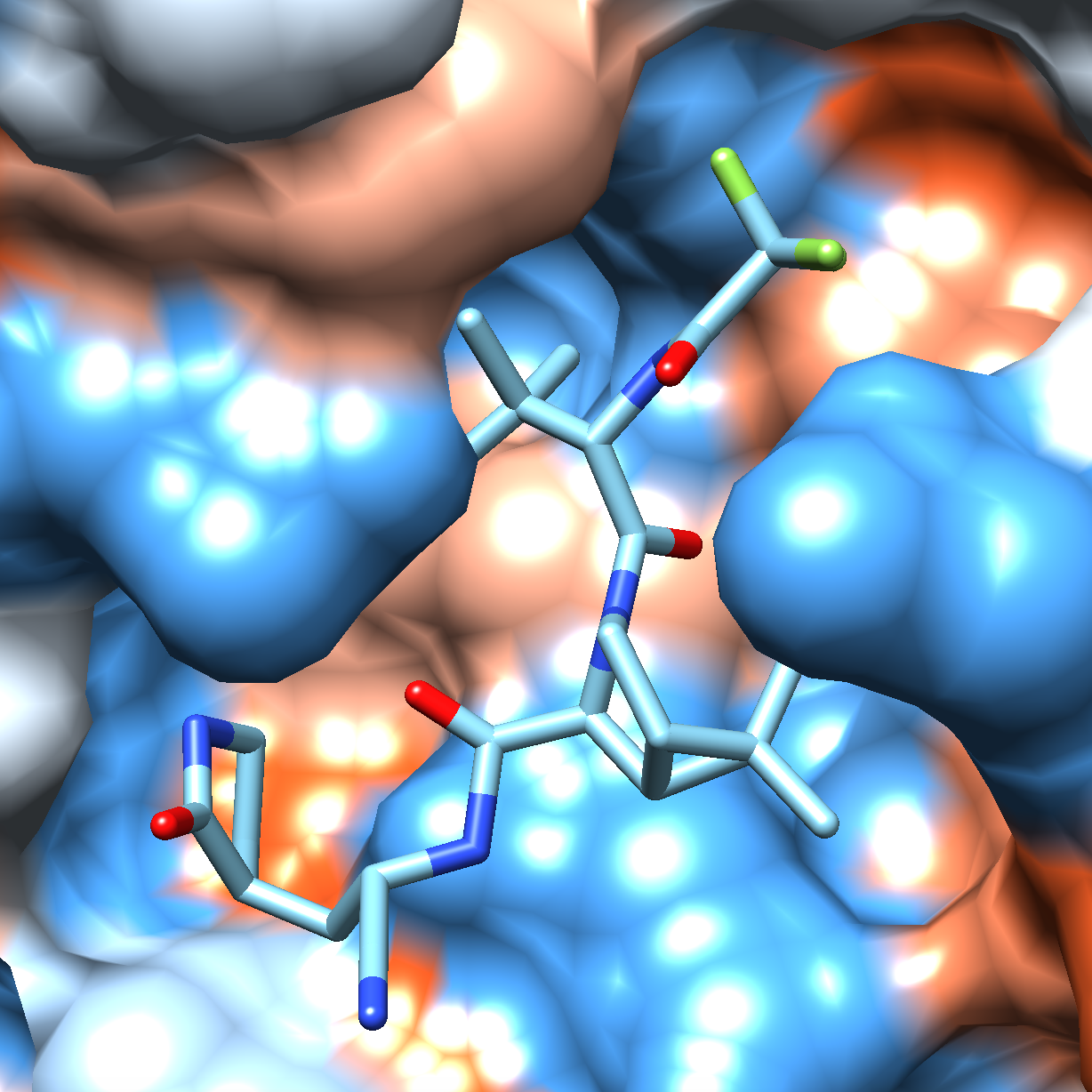}
                \caption{Hydrophobic surface view of binding site with bound ligand}
                \label{fig:sub4c}
            \end{subfigure}
        \end{tabular}
    }
    \caption{Renderings of Paxlovid}
    \label{fig:mainc}
\end{figure}

\textbf{Property Optimization}. 
Using $3CL\textsuperscript{pro}$, we also evaluated the effectiveness of the 
property optimization algorithm proposed by Schneuing et al. \cite{schneuing2023}
as described above.
To do this, we first obtain a sample with high binding affinity but low 
QED and high SAS. We find that this method of optimization is not 
very effective in our case. In particular, while we are able to increase our 
QED and SAS scores after one generation, we find that docking scores drop off 
significantly. In addition, after 5 generations, the QED and SAS have not improved 
significantly and the molecules are much more dissimilar to the original, leading to 
even worse docking scores. For example, the molecule in Figure 6a is derived 3a 
after one generation. It has a QED of 0.725, an SAS of 0.493, and an affinity score of -5.65. 
After 5 generations, we attain the molecule in Figure 7a with a QED of 0.644, an SAS of 0.395,
and an affinity score of -5.23. As a result, we did not adopt property optimization in our 
final approach.

\begin{figure}[H]
    \centering
    \resizebox{0.8\textwidth}{!}{ 
        \begin{tabular}{cc}
            \begin{subfigure}[b]{0.45\textwidth}
                \centering
                \includegraphics[width=\textwidth]{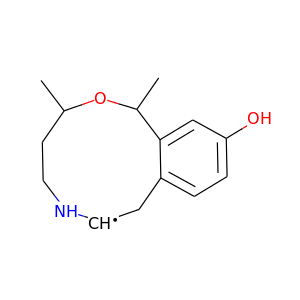}
                \caption{Illustration of molecule. This 
                molecule had a QED of 0.725 and an SAS of 0.493 and an 
                affinity score of -5.65.}
                \label{fig:sub1c}
            \end{subfigure}
            &
            \begin{subfigure}[b]{0.45\textwidth}
                \centering
                \includegraphics[width=\textwidth]{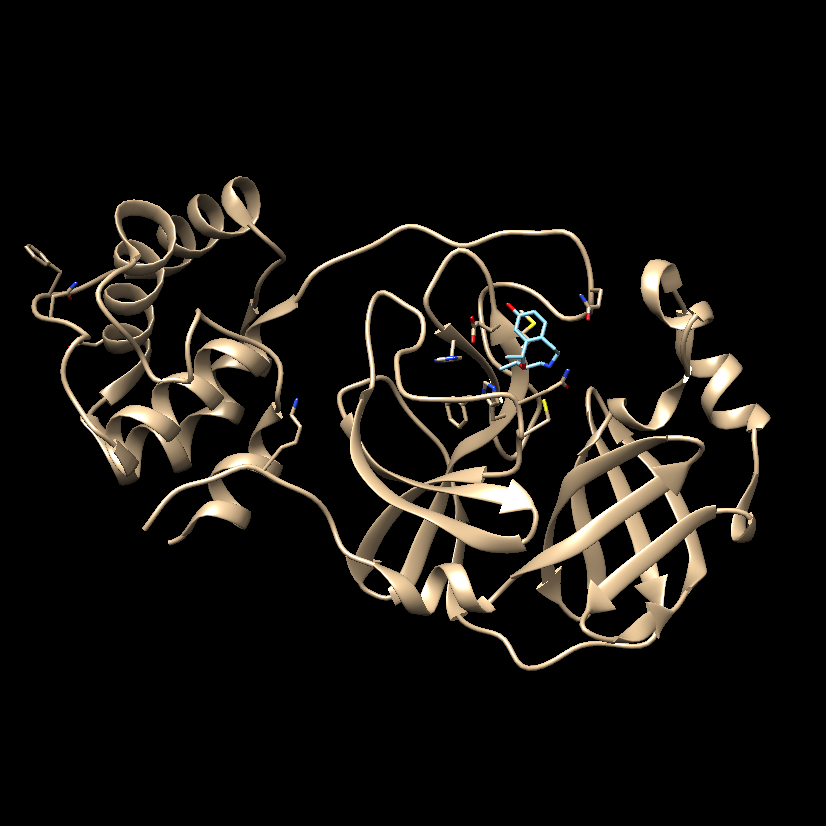}
                \caption{Rendering of molecule bound to $3CL\textsuperscript{pro}$ protease}
                \label{fig:sub2c}
            \end{subfigure}
            \\
            \begin{subfigure}[b]{0.45\textwidth}
                \centering
                \includegraphics[width=\textwidth]{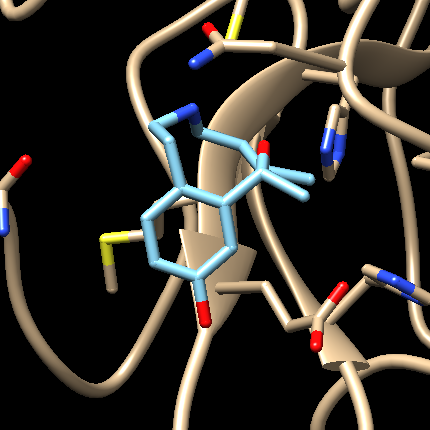}
                \caption{Rendering of molecule with $3CL\textsuperscript{pro}$ protease binding site}
                \label{fig:sub3c}
            \end{subfigure}
            &
            \begin{subfigure}[b]{0.45\textwidth}
                \centering
                \includegraphics[width=\textwidth]{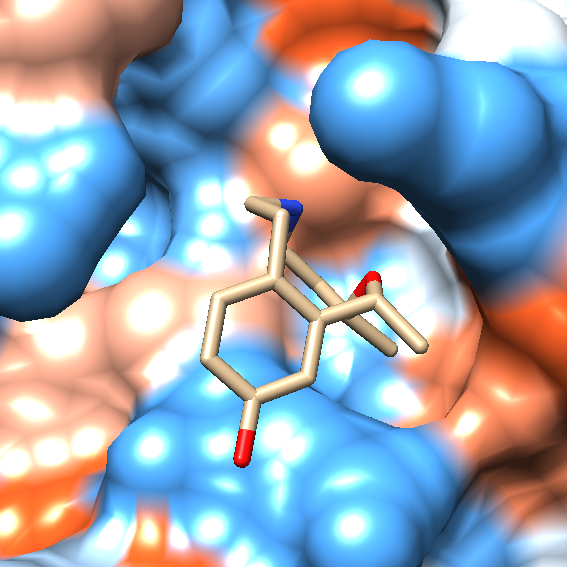}
                \caption{Hydrophobic surface view of binding site with bound ligand}
                \label{fig:sub4c}
            \end{subfigure}
        \end{tabular}
    }
    \caption{Renderings for molecule after 1 generation}
    \label{fig:mainc}
\end{figure}

\begin{figure}[H]
    \centering
    \resizebox{0.8\textwidth}{!}{
        \begin{tabular}{cc}
            \begin{subfigure}[b]{0.45\textwidth}
                \centering
                \includegraphics[width=\textwidth]{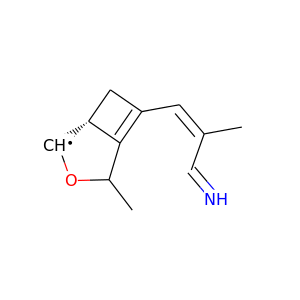}
                \caption{Illustration of molecule. This 
                molecule had a QED of 0.644 and an SAS of 0.395 and an 
                affinity score of -5.23}
                \label{fig:sub1c}
            \end{subfigure}
            &
            \begin{subfigure}[b]{0.45\textwidth}
                \centering
                \includegraphics[width=\textwidth]{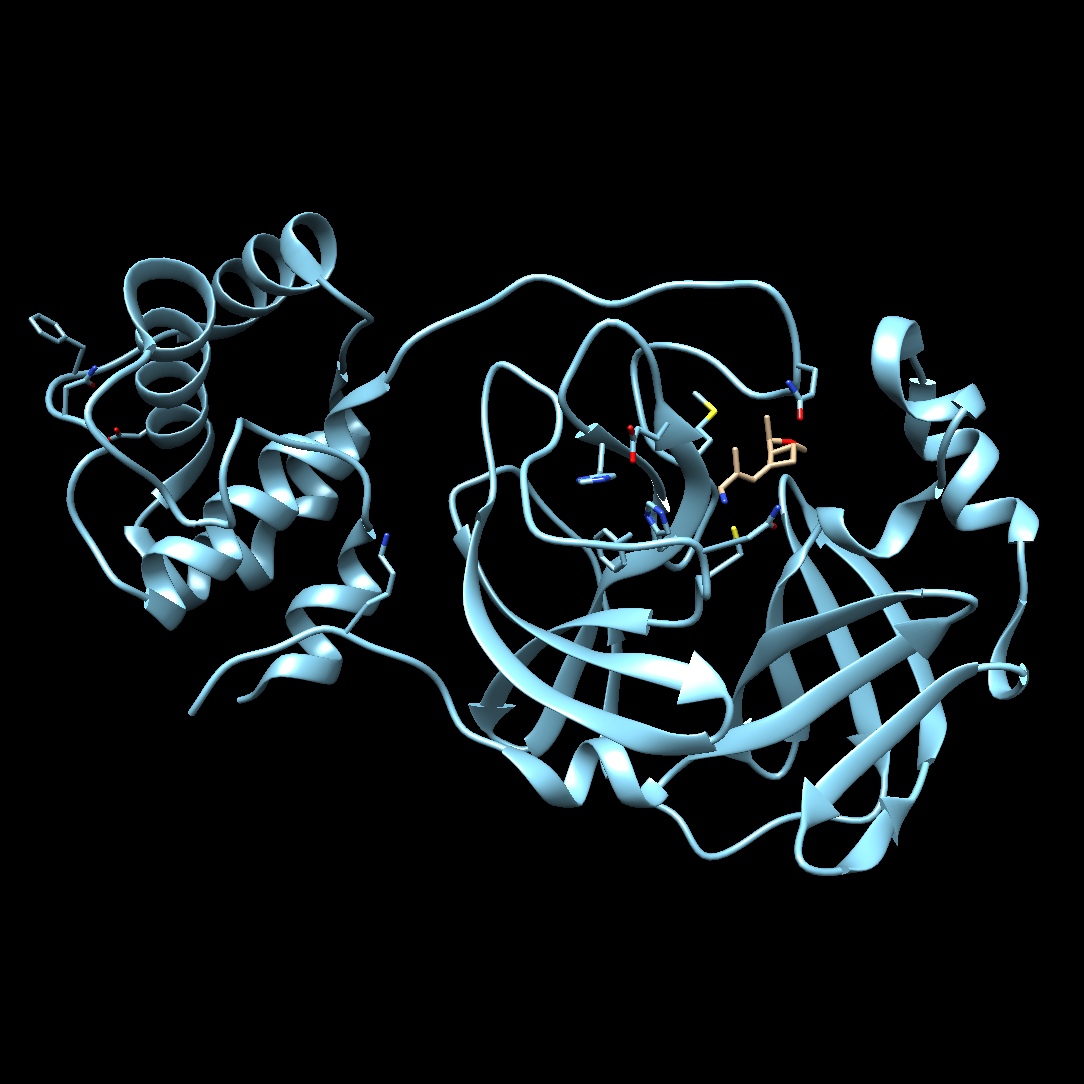}
                \caption{Rendering of molecule bound to $3CL\textsuperscript{pro}$ protease}
                \label{fig:sub2c}
            \end{subfigure}
            \\
            \begin{subfigure}[b]{0.45\textwidth}
                \centering
                \includegraphics[width=\textwidth]{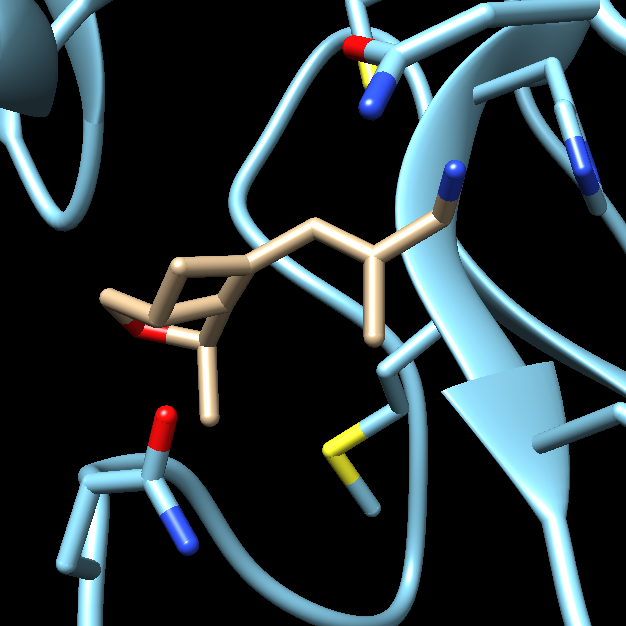}
                \caption{Rendering of molecule with $3CL\textsuperscript{pro}$ protease binding site}
                \label{fig:sub3c}
            \end{subfigure}
            &
            \begin{subfigure}[b]{0.45\textwidth}
                \centering
                \includegraphics[width=\textwidth]{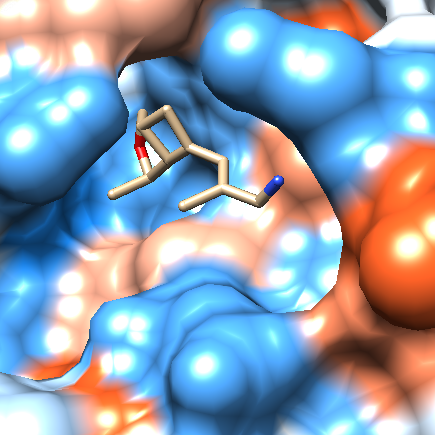}
                \caption{Hydrophobic surface view of binding site with bound ligand}
                \label{fig:sub4c}
            \end{subfigure}
        \end{tabular}
    }
    \caption{Renderings for molecule generated after 5 generations}
    \label{fig:mainc}
\end{figure}

\section{Conclusion and Future Work}

In this paper, we propose a novel method for 
\textit{de novo} drug discovery using a latent diffusion model in the 1D SMILES/SELFIES space
and demonstrated its potential to generate molecules with potency comparable to or exceeding known drugs. 
Our model stands out from existing approaches for the following reasons:
\begin{itemize}
    \item We use a novel transformer model to learn a latent representation
          of the 1D SMILES/SELFIES space (previous models have never used an 
          encoder-decoder transformer trained in a multi-task fashion for 
          molecular generation).
    \item We perform drug discovery with diffusion using 1D SMILES/SELFIES 
          (previous diffusion models were either in 2D or 3D).
    \item We use latent diffusion models for target-aware de novo drug discovery
          (previous models either did not use latent diffusion or did target-agnostic 
          drug discovery if latent diffusion was used).
\end{itemize}

Compared to other state-of-the-art models, our approach has the following advantages:
\begin{itemize}
    \item The model is able to generate molecules significantly faster than other SOTA models.
          This is because the model is using a much simpler 1D representation and 
          does not require the generation of 3D coordinates. 
          The time to generate can also be halved if accuracy is less of a concern and we perform 
          inference without classifier-free guidance.
    \item The model has the highest diversity of any SOTA model. This is because the 1D latent nature 
          of the model allows it to explore a larger space. By exploring more molecules, we can 
          find better hits, which is also why we have a relatively high Vina Top-10\% score.
    \item As we only require the amino acid sequence of the target protein, which is much easier to 
          obtain than an accurate 3D structure, we are not only able to train on a much larger dataset 
          if given more computing resources, but we can also perform inference on a much larger set of 
          proteins.
\end{itemize}

There are also a few limitations with our approach:
\begin{itemize}
    \item The model is not able to optimize properties using the method proposed by 
          Schneuing et al. \cite{schneuing2023} very well. Unfortunately, the SAS and 
          QED are somewhat random and the model quickly loses its affinity score, even with 
          only small changes in the latent space.
    \item The model has a lower average Vina score than other SOTA models. This means that 
          we must generate more molecules to find a good hit. However, since our model is 
          fast and has the highest Vina Top-10\%, we believe that this might not be a major issue.
    \item The model has relatively low average QED and SAS. This is likely connected to the fact 
          that the model explores a very large chemical space. Still, this may not be a major issue 
          since many real drugs have low QED and SAS scores and, in addition, small manual tweaks 
          can be made to the generated molecules to increase these scores.
\end{itemize}

Future work could improve our approach in the following ways:
\begin{itemize}
    \item Further optimize the model by increasing the amount of data in the training. We found that 
          the model was able to generate better molecules as we increased the training dataset size, 
          but we were limited by our resources to only train on 15,000 protein-ligand pairs. Expanding to 
          a larger portion of the PDBBind dataset could potentially improve performance.
    \item Increase the size of the diffusion model and ESM encoding. We found that the model was able to 
          generate better molecules with more parameters, but were limited by our resources. 
    \item Develop a better protein targeting approach to address the suboptimal average Vina score, enabling 
          a more efficient search process without relying solely on a large search space.
    \item Evaluate the toxicity of generated compounds by looking at potential off-target interactions. This would 
          be another step in making our molecules as realistic as possible.
\end{itemize}

\textbf{Source Code}: The source code for the project is available on GitHub.

\section{Acknowledgments}
This research was conducted entirely independently by the author.
We thank the Pingry School for generously providing us with 
the computing resources that enabled us to run our experiments.

\bibliographystyle{plain}
\bibliography{refs}

\begin{thebibliography}{10}

\bibitem{Arus-Pous2019}
Josep Ar{\ifmmode\acute{u}\else\'{u}\fi}s-Pous, Simon~Viet Johansson, Oleksii Prykhodko, Esben~Jannik Bjerrum, Christian Tyrchan, Jean-Louis Reymond, Hongming Chen, and Ola Engkvist.
\newblock {Randomized SMILES strings improve the quality of molecular generative models}.
\newblock {\em J. Cheminf.}, 11, 2019.

\bibitem{Bajusz2015Dec}
D{\ifmmode\acute{a}\else\'{a}\fi}vid Bajusz, Anita R{\ifmmode\acute{a}\else\'{a}\fi}cz, and K{\ifmmode\acute{a}\else\'{a}\fi}roly H{\ifmmode\acute{e}\else\'{e}\fi}berger.
\newblock {Why is Tanimoto index an appropriate choice for fingerprint-based similarity calculations?}
\newblock {\em J. Cheminf.}, 7(1):1--13, December 2015.

\bibitem{Bickerton2012-vd}
G~Richard Bickerton, Gaia~V Paolini, J{\'e}r{\'e}my Besnard, Sorel Muresan, and Andrew~L Hopkins.
\newblock Quantifying the chemical beauty of drugs.
\newblock {\em Nat. Chem.}, 4(2):90--98, January 2012.

\bibitem{brown2020languagemodelsfewshotlearners}
Tom~B. Brown, Benjamin Mann, Nick Ryder, Melanie Subbiah, Jared Kaplan, Prafulla Dhariwal, Arvind Neelakantan, Pranav Shyam, Girish Sastry, Amanda Askell, Sandhini Agarwal, Ariel Herbert-Voss, Gretchen Krueger, Tom Henighan, Rewon Child, Aditya Ramesh, Daniel~M. Ziegler, Jeffrey Wu, Clemens Winter, Christopher Hesse, Mark Chen, Eric Sigler, Mateusz Litwin, Scott Gray, Benjamin Chess, Jack Clark, Christopher Berner, Sam McCandlish, Alec Radford, Ilya Sutskever, and Dario Amodei.
\newblock Language models are few-shot learners, 2020.

\bibitem{biom13091339}
Andrea Citarella, Alessandro Dimasi, Davide Moi, Daniele Passarella, Angela Scala, Anna Piperno, and Nicola Micale.
\newblock Recent advances in sars-cov-2 main protease inhibitors: From nirmatrelvir to future perspectives.
\newblock {\em Biomolecules}, 13(9), 2023.

\bibitem{ERTL17}
Peter Ertl, Richard Lewis, Eric~J. Martin, and Valery~R. Polyakov.
\newblock In silico generation of novel, drug-like chemical matter using the {LSTM} neural network.
\newblock {\em CoRR}, abs/1712.07449, 2017.

\bibitem{Ertl2009}
Peter Ertl and Ansgar Schuffenhauer.
\newblock Estimation of synthetic accessibility score of drug-like molecules based on molecular complexity and fragment contributions.
\newblock {\em Journal of Cheminformatics}, 1(1):8, Jun 2009.

\bibitem{fabian2020molecularrepresentationlearninglanguage}
Benedek Fabian, Thomas Edlich, Héléna Gaspar, Marwin Segler, Joshua Meyers, Marco Fiscato, and Mohamed Ahmed.
\newblock Molecular representation learning with language models and domain-relevant auxiliary tasks, 2020.

\bibitem{gebauer20}
Niklas W.~A. Gebauer, Michael Gastegger, and Kristof~T. Schütt.
\newblock Symmetry-adapted generation of 3d point sets for the targeted discovery of molecules, 2020.

\bibitem{Bombarelli16}
Rafael G{\'{o}}mez{-}Bombarelli, David Duvenaud, Jos{\'{e}}~Miguel Hern{\'{a}}ndez{-}Lobato, Jorge Aguilera{-}Iparraguirre, Timothy~D. Hirzel, Ryan~P. Adams, and Al{\'{a}}n Aspuru{-}Guzik.
\newblock Automatic chemical design using a data-driven continuous representation of molecules.
\newblock {\em CoRR}, abs/1610.02415, 2016.

\bibitem{guan2023d}
Jiaqi Guan, Wesley~Wei Qian, Xingang Peng, Yufeng Su, Jian Peng, and Jianzhu Ma.
\newblock 3d equivariant diffusion for target-aware molecule generation and affinity prediction.
\newblock In {\em The Eleventh International Conference on Learning Representations}, 2023.

\bibitem{Guimaraes18}
Gabriel~Lima Guimaraes, Benjamin Sanchez-Lengeling, Carlos Outeiral, Pedro Luis~Cunha Farias, and Alán Aspuru-Guzik.
\newblock Objective-reinforced generative adversarial networks (organ) for sequence generation models, 2018.

\bibitem{ho2020denoisingdiffusionprobabilisticmodels}
Jonathan Ho, Ajay Jain, and Pieter Abbeel.
\newblock Denoising diffusion probabilistic models, 2020.

\bibitem{ho2022classifierfreediffusionguidance}
Jonathan Ho and Tim Salimans.
\newblock Classifier-free diffusion guidance, 2022.

\bibitem{honda2019smilestransformerpretrainedmolecular}
Shion Honda, Shoi Shi, and Hiroki~R. Ueda.
\newblock Smiles transformer: Pre-trained molecular fingerprint for low data drug discovery, 2019.

\bibitem{hoogeboom22}
Emiel Hoogeboom, Victor~Garcia Satorras, Clément Vignac, and Max Welling.
\newblock Equivariant diffusion for molecule generation in 3d, 2022.

\bibitem{kalwar2022latentganautoencoderlearningdisentangled}
Sanket Kalwar, Animikh Aich, and Tanay Dixit.
\newblock Latentgan autoencoder: Learning disentangled latent distribution, 2022.

\bibitem{kingma2017adammethodstochasticoptimization}
Diederik~P. Kingma and Jimmy Ba.
\newblock Adam: A method for stochastic optimization, 2017.

\bibitem{Krenn19}
Mario Krenn, Florian H{\"{a}}se, AkshatKumar Nigam, Pascal Friederich, and Al{\'{a}}n Aspuru{-}Guzik.
\newblock {SELFIES:} a robust representation of semantically constrained graphs with an example application in chemistry.
\newblock {\em CoRR}, abs/1905.13741, 2019.

\bibitem{rdkit}
Greg Landrum.
\newblock Rdkit: Open-source cheminformatics, 2006--.
\newblock Accessed: 2024-09-07.

\bibitem{Li23}
Yuesen Li, Chengyi Gao, Xin Song, Xiangyu Wang, Yungang Xu, and Suxia Han.
\newblock Druggpt: A gpt-based strategy for designing potential ligands targeting specific proteins.
\newblock {\em bioRxiv}, 2023.

\bibitem{lin2023dualbalancingmultitasklearning}
Baijiong Lin, Weisen Jiang, Feiyang Ye, Yu~Zhang, Pengguang Chen, Ying-Cong Chen, Shu Liu, and James~T. Kwok.
\newblock Dual-balancing for multi-task learning, 2023.

\bibitem{Lin2022.07.20.500902}
Zeming Lin, Halil Akin, Roshan Rao, Brian Hie, Zhongkai Zhu, Wenting Lu, Nikita Smetanin, Robert Verkuil, Ori Kabeli, Yaniv Shmueli, Allan dos Santos~Costa, Maryam Fazel-Zarandi, Tom Sercu, Salvatore Candido, and Alexander Rives.
\newblock Evolutionary-scale prediction of atomic level protein structure with a language model.
\newblock {\em bioRxiv}, 2022.

\bibitem{luo20223dgenerativemodelstructurebased}
Shitong Luo, Jiaqi Guan, Jianzhu Ma, and Jian Peng.
\newblock A 3d generative model for structure-based drug design, 2022.

\bibitem{nichol2021improveddenoisingdiffusionprobabilistic}
Alex Nichol and Prafulla Dhariwal.
\newblock Improved denoising diffusion probabilistic models, 2021.

\bibitem{nichol2022glidephotorealisticimagegeneration}
Alex Nichol, Prafulla Dhariwal, Aditya Ramesh, Pranav Shyam, Pamela Mishkin, Bob McGrew, Ilya Sutskever, and Mark Chen.
\newblock Glide: Towards photorealistic image generation and editing with text-guided diffusion models, 2022.

\bibitem{peng2022pocket2molefficientmolecularsampling}
Xingang Peng, Shitong Luo, Jiaqi Guan, Qi~Xie, Jian Peng, and Jianzhu Ma.
\newblock Pocket2mol: Efficient molecular sampling based on 3d protein pockets, 2022.

\bibitem{ramesh2022hierarchicaltextconditionalimagegeneration}
Aditya Ramesh, Prafulla Dhariwal, Alex Nichol, Casey Chu, and Mark Chen.
\newblock Hierarchical text-conditional image generation with clip latents, 2022.

\bibitem{ramesh2021zeroshottexttoimagegeneration}
Aditya Ramesh, Mikhail Pavlov, Gabriel Goh, Scott Gray, Chelsea Voss, Alec Radford, Mark Chen, and Ilya Sutskever.
\newblock Zero-shot text-to-image generation, 2021.

\bibitem{Reina2022-xy}
J~Reina and C~Iglesias.
\newblock Nirmatrelvir plus ritonavir (paxlovid) a potent {SARS-CoV-2} {3CLpro} protease inhibitor combination.
\newblock {\em Rev. Esp. Quimioter.}, 35(3):236--240, June 2022.

\bibitem{rombach2022highresolutionimagesynthesislatent}
Robin Rombach, Andreas Blattmann, Dominik Lorenz, Patrick Esser, and Björn Ommer.
\newblock High-resolution image synthesis with latent diffusion models, 2022.

\bibitem{schneuing2023}
Arne Schneuing, Yuanqi Du, Charles Harris, Arian Jamasb, Ilia Igashov, Weitao Du, Tom Blundell, Pietro Lió, Carla Gomes, Max Welling, Michael Bronstein, and Bruno Correia.
\newblock Structure-based drug design with equivariant diffusion models, 2023.

\bibitem{sennrich2016neuralmachinetranslationrare}
Rico Sennrich, Barry Haddow, and Alexandra Birch.
\newblock Neural machine translation of rare words with subword units, 2016.

\bibitem{sohldickstein2015deepunsupervisedlearningusing}
Jascha Sohl-Dickstein, Eric~A. Weiss, Niru Maheswaranathan, and Surya Ganguli.
\newblock Deep unsupervised learning using nonequilibrium thermodynamics, 2015.

\bibitem{vaswani2023attentionneed}
Ashish Vaswani, Noam Shazeer, Niki Parmar, Jakob Uszkoreit, Llion Jones, Aidan~N. Gomez, Lukasz Kaiser, and Illia Polosukhin.
\newblock Attention is all you need, 2023.

\bibitem{vignac2023midimixedgraph3d}
Clement Vignac, Nagham Osman, Laura Toni, and Pascal Frossard.
\newblock Midi: Mixed graph and 3d denoising diffusion for molecule generation, 2023.

\bibitem{Wang2005Jun}
Renxiao Wang, Xueliang Fang, Yipin Lu, Chao-Yie Yang, and Shaomeng Wang.
\newblock {The PDBbind database: methodologies and updates}.
\newblock {\em J. Med. Chem.}, 48(12):4111--4119, June 2005.

\bibitem{xu2023geometriclatentdiffusionmodels}
Minkai Xu, Alexander Powers, Ron Dror, Stefano Ermon, and Jure Leskovec.
\newblock Geometric latent diffusion models for 3d molecule generation, 2023.

\bibitem{Yu23}
Yuejiang Yu, Chun Cai, Jiayue Wang, Zonghua Bo, Zhengdan Zhu, and Hang Zheng.
\newblock Uni-dock: Gpu-accelerated docking enables ultralarge virtual screening.
\newblock {\em Journal of Chemical Theory and Computation}, 19(11):3336--3345, 2023.
\newblock PMID: 37125970.

\bibitem{zdrazil23}
Barbara Zdrazil, Eloy Felix, Fiona Hunter, Emma~J Manners, James Blackshaw, Sybilla Corbett, Marleen de Veij, Harris Ioannidis, David~Mendez Lopez, Juan F Mosquera, Maria Paula Magarinos, Nicolas Bosc, Ricardo Arcila, Tevfik Kizilören, Anna Gaulton, A Patrícia Bento, Melissa F Adasme, Peter Monecke, Gregory A Landrum, and Andrew R Leach.
\newblock {The ChEMBL Database in 2023: a drug discovery platform spanning multiple bioactivity data types and time periods}.
\newblock {\em Nucleic Acids Research}, 52(D1):D1180--D1192, 11 2023.

\bibitem{Zhu2020-be}
Wei Zhu, Miao Xu, Catherine~Z Chen, Hui Guo, Min Shen, Xin Hu, Paul Shinn, Carleen Klumpp-Thomas, Samuel~G Michael, and Wei Zheng.
\newblock Identification of {SARS-CoV-2} {3CL} protease inhibitors by a quantitative high-throughput screening.
\newblock {\em ACS Pharmacol. Transl. Sci.}, 3(5):1008--1016, October 2020.

\end{thebibliography}

\end{document}